\documentclass[letterpaper, 10 pt, conference]{ieeeconf}  

\IEEEoverridecommandlockouts                              
\usepackage{vaucanson-g}
\usepackage{url}

\usepackage{graphics} 
\usepackage{epsfig} 
\usepackage{multirow}
\usepackage{amsmath} 
\usepackage{amssymb}  

\title{\LARGE \bf
Sparse plus low-rank autoregressive identification in \\
neuroimaging time series* 
}

\author{Rapha\"{e}l Li\'{e}geois$^{1}$, Bamdev Mishra$^{1}$, Mattia Zorzi$^{2}$, and Rodolphe Sepulchre$^{1,3}$
\thanks{*This paper presents research results of the Belgian Network DYSCO (Dynamical Systems, Control, and Optimization), funded by the Interuniversity Attraction Poles Program, initiated by the Science Policy Office of the Belgian State. The scientific responsibility rests with its authors. BM is supported as an FRS-FNRS research fellow (Belgian Fund for Scientific Research).}%
\thanks{$^{1}$RL, BM and RS are with the Department of Electrical Engineering and Computer Science, University of Li\`{e}ge, Li\`{e}ge, Belgium.
        {\tt\small \{R.Liegeois,B.Mishra\}@ulg.ac.be}}%
\thanks{$^{2}$MZ is with the Department of Information Engineering, University of Padova, 35131 Padova, Italy.
        {\tt\small ZorziMat@dei.unipd.it}}%
\thanks{$^{3}$RS is with the Department of Engineering, Trumpington Street, University of Cambridge, Cambridge CB2 1PZ, United Kingdom.
        {\tt\small R.Sepulchre@eng.cam.ac.uk}}%
        }

\begin{document}
\maketitle
\thispagestyle{empty}
\pagestyle{empty}

\begin{abstract}

This paper considers the problem of identifying multivariate autoregressive (AR) sparse plus low-rank graphical models. Based on the corresponding problem formulation recently presented, we use the alternating direction method of multipliers (ADMM) to efficiently solve it and scale it to sizes encountered in neuroimaging applications. We apply this decomposition on synthetic and real neuroimaging datasets with a specific focus on the information encoded in the low-rank structure of our model. In particular, we illustrate that this information captures the \emph{spatio-temporal structure} of the original data, generalizing classical component analysis approaches.  
\end{abstract}

\section{Introduction}
Identifying the links between the variables in multivariate datasets is a fundamental and recurrent problem in many engineering applications. To this end, the use of graphs especially in neuroimaging applications has become very popular because they allow to study and represent the interactions between variables in a concise manner \cite{bullmore}. 

A particular class of graphs, called \emph{graphical models}, encodes information about dependence between the variables conditioned on all the other variables, or \emph{conditional dependence}  \cite{koller}. For \emph{static} models this information is contained in the inverse of the covariance matrix also known as the \emph{precision} matrix, and can be estimated by solving the covariance selection problem \cite{dempster}. Additionally, sparsity and/or low-rank structural constraints can be imposed to the precision matrix estimation. The sparsity constraint results from the parsimony principle in model fitting, i.e., one assumes few direct interactions between the variables, and is obtained through $l1$-norm regularizers \cite{daspremont}. The low-rank structure, induced through nuclear-norm regularizers, models the presence of \emph{latent variables} that are not observed but generate a common behavior in all the observed variables \cite{chandrasekaran}. The low-rank modeling is inspired from what is done in classical component analysis techniques, and leads to models that are simpler and more interpretable \cite{pca,ica}. An example of graphical model is given in Fig. 1


         \begin{figure}[thpb]
      \centering
     
      \includegraphics[width=0.5\textwidth]{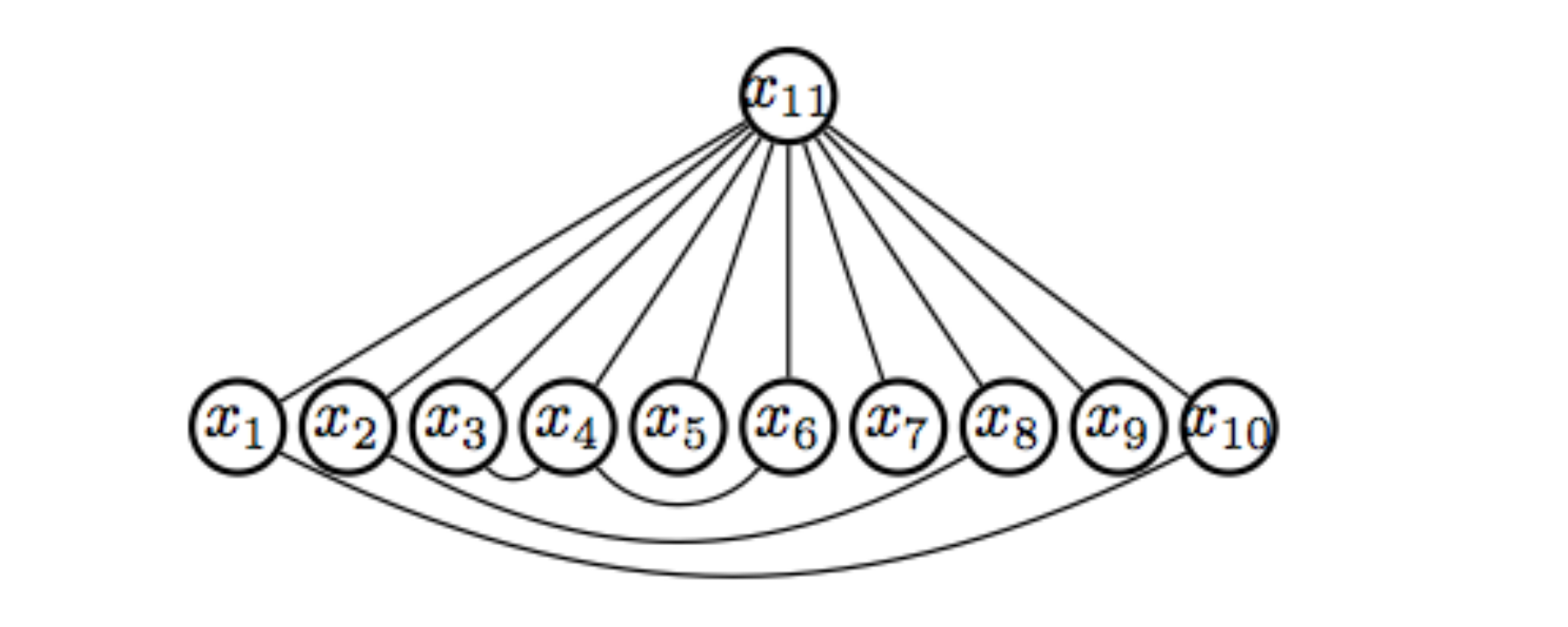}
      \caption{Ten observed variables $x_i, i\in[1 \ldots 10]$ with few interactions among them (sparsity) and one latent variable $x_{11}$ (low-rank structure).}
      \label{interaction}
   \end{figure}

In \emph{dynamical} models, the additional information of the ordering of the data is taken into account and datasets are seen as time series. A widely used class of models encoding this information are autoregressive (AR) models which are characterized by their power spectral density, the dynamic equivalent of the covariance matrix \cite{stoica}. As in the static case, it has been shown that a zero in the inverse power spectral density corresponds to conditional independence between two variables \cite{brillinger}. In the dynamic case the (inverse) power spectral density is encoded in a block Toeplitz matrix. Because of this particular structure the classical $l1$-norm can not be used to induce sparsity in the inverse power spectral density. This problem is solved by introducing an alternate regularization proposed in \cite{songsiri}. Finally, \cite{zorzi} presents a unifying framework allowing sparse plus low-rank identification of inverse power spectral densities in multivariate time series.


In this paper we adapt the problem formulated in \cite{zorzi} to the alternating direction method of multipliers (ADMM) framework of \cite{boyd} in order to scale it with larger datasets for which the CVX Matlab toolbox of \cite{grant} is computationally expensive. In particular, we exploit separability of constraints of the ADMM framework to decouple the sparsity and the low-rank constraints. The first update is a projective gradient update similar to the one proposed in \cite{songsiri} and the second update is a well known projection onto the cone of positive semidefinite matrices. In the numerical examples we show the efficacy of our proposed algorithm on a real neuroimaging dataset. We  also provide further insight into the information encoded in the low-rank structure of our model by applying the proposed algorithm to datasets with different spatio-temporal structures, which is shown to be at least partially recovered in the latent variables.

The paper is organized as follows. We present the optimization problem leading to this sparse plus low-rank decomposition in Section II and we explain how we use ADMM to efficiently solve it in Section III. We then show the results of our approach on synthetic and real data in Section IV and conclude.

\section{Problem formulation}

We first introduce some basic notions, explain the motivation of the sparse plus low-rank (S+L) graphical models and then formally deduce the corresponding optimization problem. Finally, we define \emph{latent components} that we use in the numerical examples in order to characterize information encoded in the low-rank part of this decomposition.

Consider a $q$-dimensional autoregressive (AR) gaussian process $x=[x_1(t) \ldots x_q(t)]^T$ of order $p$
\begin{eqnarray*}
x(t)=\sum_{i=1}^p A_i x(t-i) + \epsilon(t),
\end{eqnarray*}
where $x(t) \in \mathbb{R}^q$, $A_i \in \mathbb{R}^{q \times q}$, $i=1 \ldots p$ and $\epsilon(t)$ is white gaussian noise with covariance matrix $\Sigma$. $x$ is completely characterized by its spectral density $\Phi(e^{j \theta})$ which encodes information about dependence relations between the $q$ variables \cite{stoica}. On the contrary the \emph{inverse} power spectral density $\Phi_x(e^{j \theta})^{-1}$ encodes \emph{conditional} dependence relations between variables \cite{brillinger,dahlhaus}. That is, two variables $x_k$ and $x_l$ are independent, conditionally on the other $q-2$ variables of $x$ over $\{t\in \mathbb{Z}\}$, if and only if 
\begin{eqnarray}
(\Phi(e^{j \theta})^{-1})_{kl} = 0 \ \ \ \forall \theta \in [0,2\pi],
\label{eq:sparsity}
\end{eqnarray}
The nodes of the corresponding graphical model are the $q$ variables of $x$ and there is no edge between the two nodes $x_k$ and $x_l$ if and only if (\ref{eq:sparsity}) is satisfied. 

\subsection{S+L graphical models}

Assume that $x(t)=[(x^m(t))^T \ (x^l(t))^T]^T$ where $x^m(t)=[x_1(t) \ldots x_m(t)]^T \in \mathbb{R}^m$ contains \emph{manifest} variables, that is variables accessible to observations, and $x^l(t)=[x_{m+1}(t) \ldots x_{m+l}(t)]^T \in \mathbb{R}^l$ contains \emph{latent} variables, not accessible to observations. The power spectrum of $x$ can be expressed using the following block decomposition

\begin{eqnarray}
\Phi_x = \begin{bmatrix} 
\Phi_m & \Phi_{ml}^* \\
\Phi_{ml} & \Phi_l 
\end{bmatrix},\ \ \
\Phi_x^{-1} = \begin{bmatrix} 
\Upsilon_m & \Upsilon_{ml}^* \\
\Upsilon_{ml} & \Upsilon_l 
\end{bmatrix},
\label{eq:block}
\end{eqnarray}
where the $^*$ denotes the conjugate transpose operation.

In order to better characterize the conditional dependence relations between the manifest variables, from (\ref{eq:block}) we obtain the following decomposition of $\Phi_m^{-1}$ using the Schur complement \cite{horn}
\begin{eqnarray}
\Phi_m^{-1} = \Upsilon_m -  \Upsilon_{ml}^* \Upsilon_{l} \Upsilon_{ml}\ .
\label{eq:decsl}
\end{eqnarray}

The main modeling assumption here is that $l \ll m$ and that the conditional dependencies relations among the $m$ manifest variables encoded in $\Phi_m^{-1}$ can be largely explained through few latent variables. The corresponding graphical model has few edges between the manifest variables and few latent nodes, as in Fig. 1. This leads to a S+L structure for $\Phi_m^{-1}$ following (\ref{eq:decsl}): $\Phi_m^{-1} = \Sigma -  \Lambda$, where $\Sigma=\Upsilon_m$ is sparse because it encodes conditional dependence relations between the manifest variables and $\Lambda=\Upsilon_{ml}^* \Upsilon_{l} \Upsilon_{ml}$ is low rank. Since $x$ is an AR process of order $p$, we can assume that $\Sigma$ and $\Lambda$ belong to the family of matrix pseudo-polynomials 
\begin{eqnarray*}
\mathcal{Q}_{m,p}=\{ \sum_{j=-p}^p e^{-ij \theta} R_j \text{ : } R_j=R^T_{-j} \in \mathbb{R}^{m \times m} \}.
\end{eqnarray*}

Following \cite{zorzi} we further rewrite $\Sigma$ and $\Lambda$ as
\begin{eqnarray}
\begin{tabular}{rcl}
              $\Sigma - \Lambda$ &$=$& $\Delta X \Delta^*$,\\
              $\Lambda $&$=$&$ \Delta L \Delta ^*$,             
\end{tabular}
\label{eq:sl}
\end{eqnarray}
where $\Delta$ is a shift operator $\Delta(e^{i\theta}):=[I \text{ \hspace{0.2cm}} e^{i\theta}I \text{ } \ldots \text{ } e^{ip\theta}I]$ and $X$ and $L$ are now matrices belonging to $Q_{m(p+1)}$ which is the set of symmetric matrices of size $m(p+1) \times m(p+1)$.

Finally, $M_{m,p}$ is the vector space of matrices $W:=[W_0 \text{ } W_1 \ldots W_p]$ with $W_0 \in Q_m$ and $W_1 \ldots W_p \in \mathbb{R}^{m \times m}$. The linear mapping $\mathcal{T}: M_{m,p} \rightarrow Q_{m(p+1)}$ outputs a symmetric block Toeplitz matrix from the blocks of $W$ as
\begin{eqnarray*}
\mathcal{T}(W) =
 \begin{pmatrix}
  W_0 & W_1 & \cdots & W_p \\
W_1^T& W_0 & \ddots & \vdots \\
  \vdots  & \ddots  & \ddots & W_1  \\
  W_n^T & \cdots & W_1^T & W_0
 \end{pmatrix}.
\end{eqnarray*}

The adjoint operator of $\mathcal{T}$ is the linear mapping $\mathcal{D}: Q_{m(p+1)}\rightarrow M_{m,p}$ defined for a matrix $X \in Q_{m(p+1)}$ partitioned in square blocks of size $m\times m$ as
\begin{eqnarray}
X =
 \begin{pmatrix}
  X_{00} & X_{01} & \cdots & X_{0p} \\
X_{00}^T& X_{11} & \cdots & X_{1p} \\
  \vdots  & \vdots  & \ddots & \vdots  \\
  X_{0p}^T & X_{1p}^T & \cdots & X_{pp}
 \end{pmatrix}.
 \label{eqx}
\end{eqnarray}

Following this partition, $W=\mathcal{D}(X) \in M_{m,p}$ is given by
\begin{eqnarray}
\begin{tabular}{rrcl}
              \multirow{2}{.1cm}{$\left\{\rule{0mm}{5mm}\right.$}&$W_0(X)$ &$=$& $\sum_{h=0}^p X_{hh},$\\
              &$W_j(X)$&$=$&$ 2 \sum_{h=0}^{p-j} X_{h, h+j} \text{\hspace{.3cm}} \forall j=1 \ldots p$.             
\end{tabular}
\label{eq:D}
\end{eqnarray}

\subsection{S+L identification problem formulation}
Assume that we have a finite length realization $x^m(1) \ldots x^m(N)$ of the manifest process $x^m$. It should be emphasized that no data is available regarding the latent process $x^l$, and its dimension is not even known. Our goal is to recover the S+L model defined in the previous section which best explains the collected data. Therefore, one estimate of $X$ and $L$ (hence of $\Sigma$ and $\Lambda$) is given by solving the regularized maximum entropy problem \cite{byrnes,cover} for which the primal is
\begin{eqnarray}
\begin{tabular}{cl}
              $\displaystyle\min_{X,L \in Q_{m(p+1)}}$&$- \log \det X_{00} + \langle C,X \rangle $\\
              & $\text{\hspace{.5cm}}+ \lambda \gamma h(\mathcal{D}(X+L)) + \lambda \text{ } \text{trace}(L)$              \vspace{.2cm}\\
              subject to& $X \succeq 0, X_{00} \succ 0, L \succeq 0$,
         \end{tabular}
         \label{primal}
\end{eqnarray}
  where
  \begin{itemize}
  \item $\gamma>0$ and $\lambda>0$ are weighting parameters leading to a sparse $\Sigma$ and a low-rank $\Lambda$,
  \item $C=\mathcal{T}(\hat{R})$ where $\hat{R} \in M_{m,p}$ are the $p+1$ first sample covariance lags $\hat{R}_k$ \cite{stoica},
  \item $\langle .,.\rangle$ is the inner product associated with $\mathbb{R}^{m \times m}$,
  \item $h$ is the following function chosen to encourage a structured sparse solution $\mathcal{D}(X+L)$
  \end{itemize} 
$$h(Y)=\sum_{j>i} \max \{|(Y_0)_{ij}|, \max_{k=1,...,p}|(Y_k)_{ij}|,\max_{k=1,...,p}|(Y_k)_{ji}|\},$$
which is convex but non smooth. This limitation was contoured in similar works \cite{songsiri} by solving the corresponding dual problem. In the present case, it can be shown based on \cite{songsiri} and \cite{zorzi} that the dual of (\ref{primal}) is

\begin{eqnarray}
\begin{tabular}{cl}
              $\displaystyle\max_{Z \in{M_{m,p}}}$&$ \Psi(C+\mathcal{T}(Z))$\\
 subject to&$\sum_{k=0}^p (|(Z_k)_{ij}|+|(Z_k)_{ji}|)\leq \gamma \lambda, \text{ } i \neq j$\vspace{0.2cm}\\
 &$\textbf{diag}(Z_k)=0, \text{ } k=0,...,p$\vspace{0.2cm}\\
 &${\lambda I + \mathcal{T}(Z) \succeq 0},$
    \end{tabular}
    \label{dual}
\end{eqnarray}
  where $\Psi : Q_{m(p+1)} \rightarrow \mathbb{R}$ is defined as
\begin{eqnarray}
\Psi(V)= - \log \det (V_{00} - V^T_{1:p,0}V^{-1}_{1:p,1:p}V_{1:p,0}) - m  \nonumber
\end{eqnarray}
and $V \in Q_{m(p+1)}$ is partitioned as in (\ref{eqx}). 

\subsection{Definition of latent components}
\label{lat_comp}
The optimal primal variables ($X^{opt}$,$L^{opt}$) are recovered from the optimal solution $Z^{opt}$ following \cite{zorzi} from which $\Sigma^{opt}$ and $\Lambda^{opt}$ are computed by (\ref{eq:sl}). $\Lambda^{opt}$ is a square matrix function of size $m$, defined over the unit circle and for which we consider the pointwise singular value decomposition since $\Lambda^{opt}$ is low-rank for all $\theta \in [0 , 2\pi]$
\begin{eqnarray}
\Lambda^{opt}_{m,m} (\theta)=\Gamma_{m,l}(\theta) \Omega_{l,l}(\theta) \Gamma_{m,l}^*(\theta).
\label{svd}
\end{eqnarray}

The $i$-th column of $\Gamma_{m,l}(\theta)$ contains the strength of the conditional dependence relation between the $i$-th unobserved latent variable and each of the $m$ manifest variables. Following component analysis nomenclature \cite{pca} we call the $i$-th column of $\Gamma_{m,l}(\theta)$ the $i$-th \emph{latent component} and denote it $\gamma_{i,.}(\theta)$. In other words, $\gamma_{i,j}(\theta)$ with $i \in [1 \ldots l]$ and $j \in [1 \ldots m]$ represents the weight of the conditional dependence between the latent variable $x_{m+i}(t)$ and the manifest variable $x_j(t)$ in the expression of $x_j(t)$. In the static case, $\gamma_{i,.}(\theta)$ reduces to a constant vector because $\Delta=I$ in (\ref{eq:sl}) and $\Lambda$ does not depend on $\theta$. On the contrary, in the dynamic case each latent component is a function of $\theta \in [0 , 2\pi]$.

\section{Alternating direction method of multipliers}
We use the alternating direction method of multipliers (ADMM) of \cite{boyd} to solve (\ref{dual}). This choice is motivated by the fact that this algorithm both inherits the strong convergence properties of the method of multipliers and exploits decomposability of the dual problem formulation leading to efficient partial updates of the variables. We show how we rewrite (\ref{dual}) in the ADMM format by separating the sparse and low-rank constraints, then explain how we choose an adequate stopping criterion and recover the primal variables.

In order to decouple the constraints related to sparsity and low-rank we introduce the new variable $Y\in Q_{m(p+1)}$ and reformulate (\ref{dual}) as
\begin{eqnarray}
\begin{tabular}{rlc}
              $\displaystyle\min_{Z\in M_{m,p}}$&$ \Psi(C+\mathcal{T}(Z))$&\\
subject to &$A(Z)\leq b$, & ($C_1$)\\
 &$Y=\mathcal{T}(Z)+\lambda I$, & \\
 &$Y \succeq 0$& ($C_2$),
         \end{tabular}
         \label{opt}
\end{eqnarray}
where the first constraint ($C_1$) gathers the first two constraints on $Z$ of (\ref{dual}) and $A$ and $b$ are defined accordingly; and ($C_2$) is the last constraint of (\ref{dual}) imposing positive semidefiniteness of the new variable $Y$. Using the \emph{augmented Lagrangian} formulation, we introduce $L_{\rho}$ defined by
\begin{eqnarray*}
L_{\rho}(Z,Y,M)=\Psi(C+T(Z))-\langle M,Y-\mathcal{T}(Z)-\lambda I \rangle \\
+ \frac{\rho}{2}\|Y-\mathcal{T}(Z)-\lambda I \|^2_F, 
\end{eqnarray*}
where $\| \cdot \|_F$ is the Frobenius norm. Subsequently, the ADMM updates are \vspace{-.3cm}
\begin{eqnarray*}
\begin{tabular}{rrll}
		\multirow{6}{.1cm}{$\left\{\rule{0mm}{10mm}\right.$}&&&\vspace{.05cm}\\
              &$Z^{k+1}=$& $\min \text{ } L_{\rho}(Z,Y^k,M^k),$&\hspace{.3cm}(S1)\vspace{-.15cm}\\
              &&\hspace{-.12cm}\scriptsize{$Z \in C_1$}&\vspace{-.05cm}\\
              &$Y^{k+1}=$&$\min \text{ } L_{\rho}(Z^{k+1},Y,M^k),$&\hspace{.3cm}(S2)\vspace{-.15cm}\\  
              &&\hspace{-.12cm}\scriptsize{$Y \in C_2$}&\vspace{-.05cm}\\
              &$M^{k+1}=$&$M^k -\rho (Y^{k+1}-\mathcal{T}(Z^{k+1})-\lambda I).$&\hspace{.3cm}(S3)           
\end{tabular}
\end{eqnarray*}

It should be noted that (S1) has no closed form solution and corresponds to the sparsity set of constraints of \cite{songsiri}. We approximate the solution by a projective gradient step as in \cite{songsiri}. Following this approach,
\begin{eqnarray*}
Z^{k+1}=\Pi_{C_1}(Z^k - t_k \nabla_Z L_{\rho}(Z_k, Y_k, M_k)),
\end{eqnarray*}
where
\begin{itemize}
\item $\nabla_Z L_{\rho}(Z_k, Y_k, M_k) = \nabla \Psi(Z^k) + \mathcal{D}(M^k)$
\item[] \hspace{4.3cm} $+ \rho \mathcal{D}(\mathcal{T}(Z^k)+\lambda I - Y^k),$
\item $t_k$ is found from the Armijo conditions,
\item $\Pi_{C_1}$ is the projection onto $C_1$ which reduces to a projection onto the $l_1$-norm ball \cite{songsiri,berg}. 
\end{itemize}

The optimization problem (S2) has a closed form solution and is computed as
\begin{eqnarray*}
Y^{k+1}=\Pi_{C_2}(\frac{1}{\rho}M^k-\mathcal{T}(Z^{k+1})-\lambda I),
\end{eqnarray*}
where $\Pi_{C_2}$ is the projection onto the cone of symmetric positive semidefinite matrices of size $m(p+1) \times m(p+1)$, which is done by selecting the eigenvectors corresponding to \emph{positive} eigenvalues. This leads to the final updates of the ADMM algorithm.\\
\noindent
\rule{8.6cm}{.8pt}
\textbf{ADMM for sparse plus low-rank inverse power spectral density estimation.} Initialize $Z_0$, $Y_0$, $M_0$; set $\rho > 0$; and successively update variables as follows:\\
\vspace{-.5cm}
\begin{eqnarray}
\begin{tabular}{rl}
$Z^{k+1}=$&$\Pi_{C_1}(Z^k - t_k \nabla_Z L_{\rho}(Z_k, Y_k, M_k)),$\\
$Y^{k+1}=$&$\Pi_{C_2}(\frac{1}{\rho}M^k-\mathcal{T}(Z^{k+1})-\lambda I),$\\
$M^{k+1}=$&$M^k -\rho (Y^{k+1}-\mathcal{T}(Z^{k+1})-\lambda I).$
\end{tabular}
\label{admm_update}
\end{eqnarray}
\rule{8.6cm}{.8pt}

Following \cite{boyd}, a stopping criterion for (\ref{admm_update}) is based on the primal and dual \emph{residuals} $r$ and $s$ that respectively measure satisfaction of the equality constraint of (\ref{opt}) and the distance between two successive iterates of the additional variable $Y$. $r$ and $s$ should satisfy $\|r\|_F \leq \epsilon^{pri}$ and $\|s\|_F \leq \epsilon^{dual}$ where $\epsilon^{pri}$ and $\epsilon^{dual}$ are defined as
\begin{eqnarray*}
\epsilon^{pri}&=&m(p+1) \epsilon^{abs} \\
&& \hspace{0.2cm} +\epsilon^{rel} \max \{ \lambda \sqrt{m(p+1)}, \|T(Z^k)\|_F,\|Y^k\|_F\}, \\
\epsilon^{dual}&=&m \sqrt{p+1} \epsilon^{abs}+\epsilon^{rel} \|\mathcal{D}(M^k)\|_F.
\end{eqnarray*}

Here $\epsilon^{abs}$ and $\epsilon^{rel}$ are the predefined absolute and relative tolerances for the problem.

A variation is obtained when $\rho$ is multiplied by a factor of $\tau > 1$ at each iteration up to a maximum value $\rho_{max}$ starting from a value $\rho_0$ depending on the application.

Convergence analysis of the ADMM algorithm follows from \cite[Section 3.2]{boyd}\cite{ouyang}. The computational cost per iteration of the updates (\ref{admm_update}) depends on the projections onto $C_1$ and $C_2$, the gradient evaluation of $L_{\rho}$, and the linear mappings $\mathcal{T}$ and $\mathcal{D}$ leading to a final complexity $\mathcal{O}(m^3 (p+1)^3)$.

\section{Numerical examples}

In this section we apply the proposed ADMM algorithm to solve the sparse plus low-rank decomposition on synthetic and real datasets and explore the type of information encoded in the identified latent \emph{components}. The Matlab code for the algorithm is available from the webpage \url{http://www.montefiore.ulg.ac.be/~rliegeois/}
\subsection{Application on linear synthetic data}
\begin{figure}[thpb]
      \centering
     
      \includegraphics[width=0.5\textwidth]{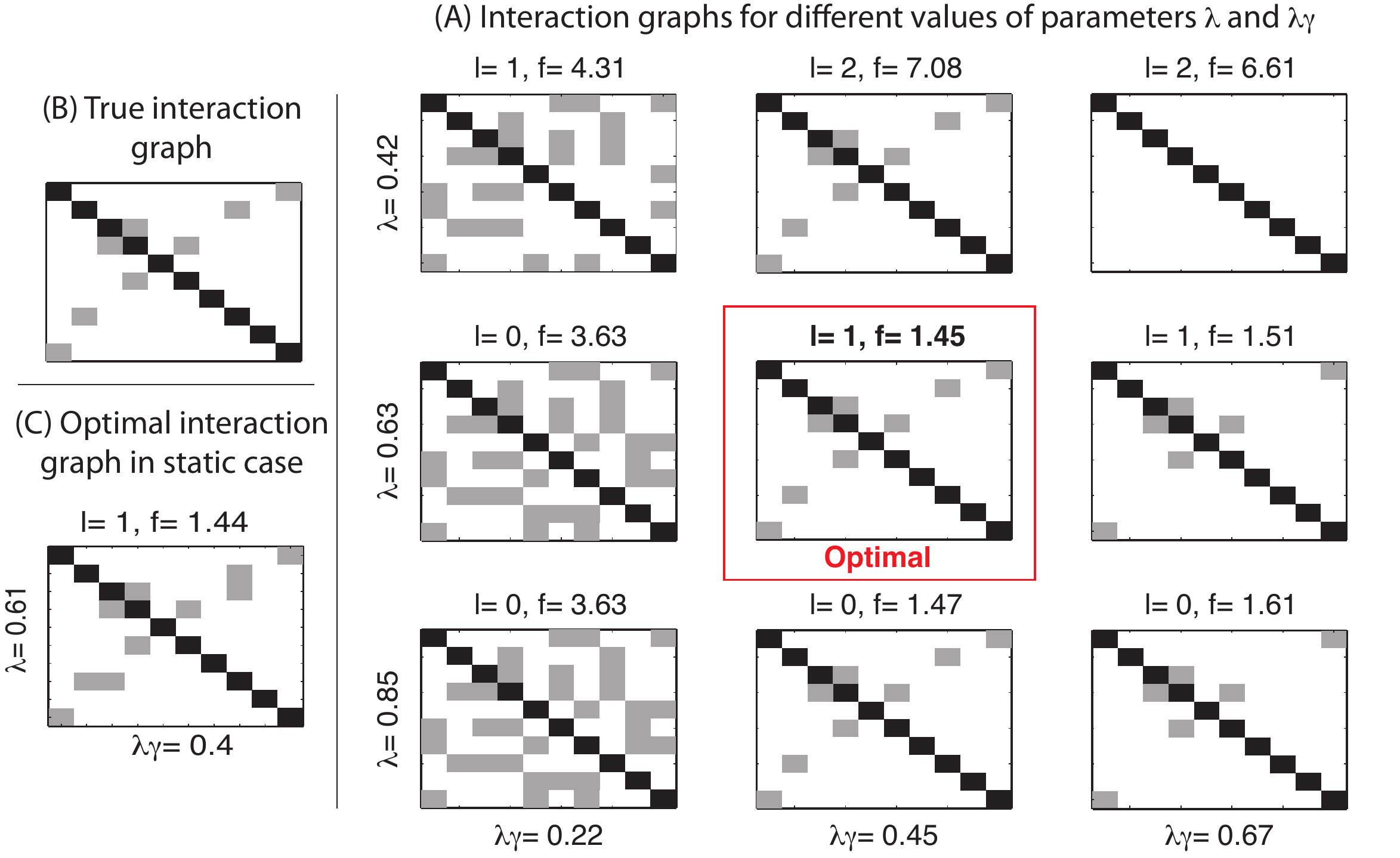}
      \caption{(A) Interaction graphs of estimated models for different values of $\lambda$ and $\lambda \gamma$ using a first order AR model ($p=1$). (B) True interaction graph. (C) Optimal interaction graph of estimated model obtained in the static case ($p=0$).}
      \label{ex1}
   \end{figure}

This synthetic dataset consists of time series corresponding to a first order AR model (dynamic model, $p=1$) with the interaction graph presented in Fig. \ref{interaction}. The interaction graphs of the manifest variables (support of $\Sigma$) identified for different values of $\lambda$ and $\lambda \gamma$ are represented in Fig. \ref{ex1} as well as $l$, the number of latent components (rank of $\Lambda$) that were identified. In order to discriminate between models we compute a score function $f$, defined in \cite{songsiri}, taking into account fitting to the data and complexity of the model. 
   
As expected, higher values of $\lambda$ promote models with less latent components, and higher values of $\lambda \gamma$ favor models with few interactions between the manifest variables. The model with the best (lowest) score function recovers the true interaction graph with the correct number of latent components (Fig. \ref{ex1}A). The optimal interaction graph in the static case (Fig. \ref{ex1}C), on the contrary, does not recover exactly the true interaction graph.

\begin{figure}[thpb]
      \centering     
      \includegraphics[width=0.5\textwidth]{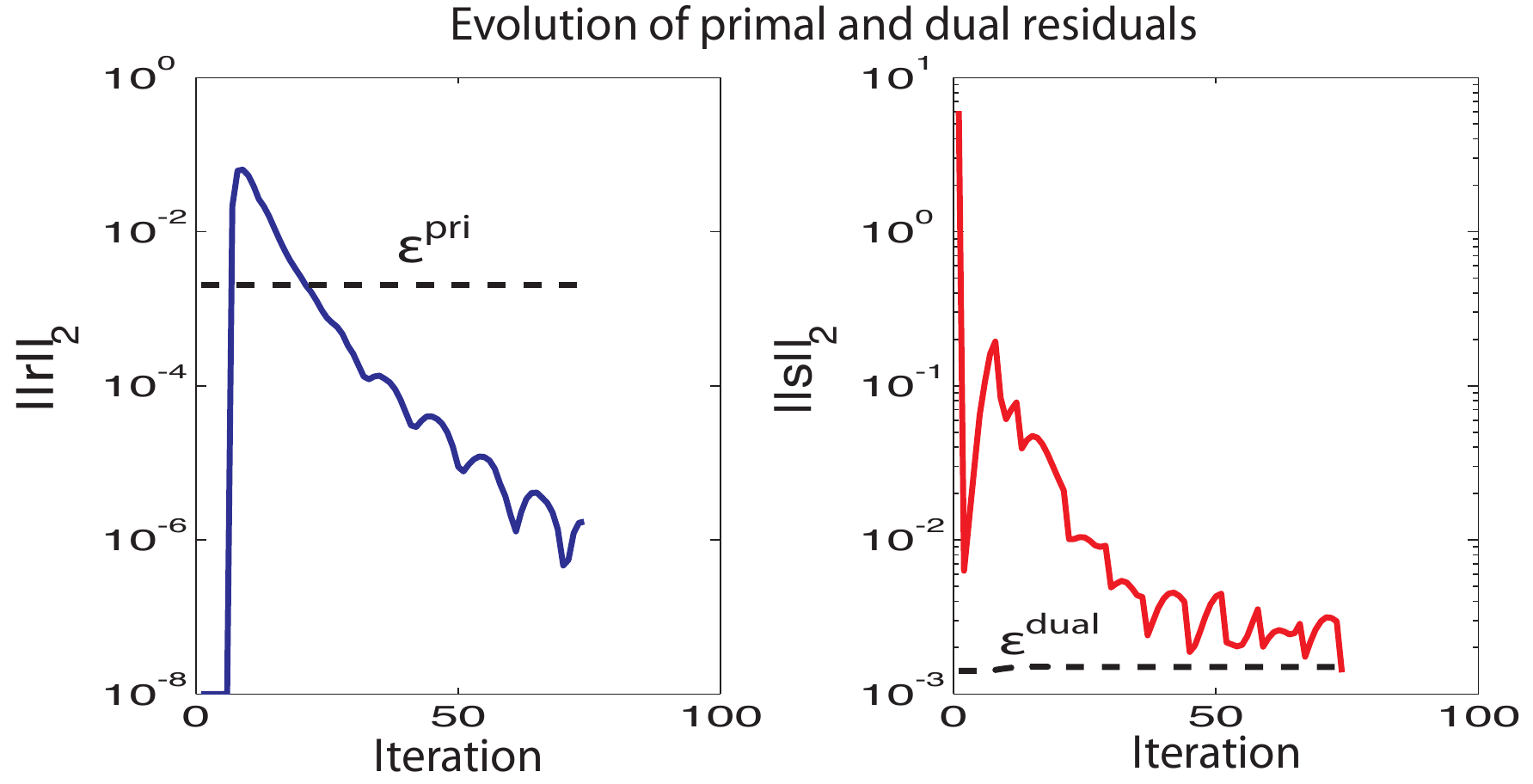}
      \caption{Typical convergence scheme of the ADMM algorithm obtained with the linear synthetic dataset with the parameter values $\rho_0=1$, $\tau=1.1$, $\rho_{max}=1000$, $\epsilon^{abs}=10^{-5}$, and $\epsilon^{del}=10^{-4}$.}
      \label{conv}
   \end{figure}

The stopping criterion based on the primal and dual residuals is illustrated in Fig. \ref{conv} using this dataset. The algorithm stops when $\|r\|_2 \leq \epsilon^{pri}$ and $\|s\|_2 \leq \epsilon^{dual}$ are both satisfied.
   
\vspace{-.4cm}
\subsection{Application on non-linear synthetic data}
The interest in considering a non-linear generative model is that it can produce endogenous sustained oscillations in networks similarly to what is observed in neuroimaging data such as fMRI time series. A popular non-linear model is the Hopfield model widely used in neural networks \cite{hopfield}: 
\begin{eqnarray*}
\dot{x_i}=-x_i+\text{sat} (\sum_{j \neq i} G_{ij} x_j) +\epsilon,
\label{Hopfield}
\end{eqnarray*}

   \begin{figure*}[thpb]
      \centering     
      \includegraphics[width=1\textwidth]{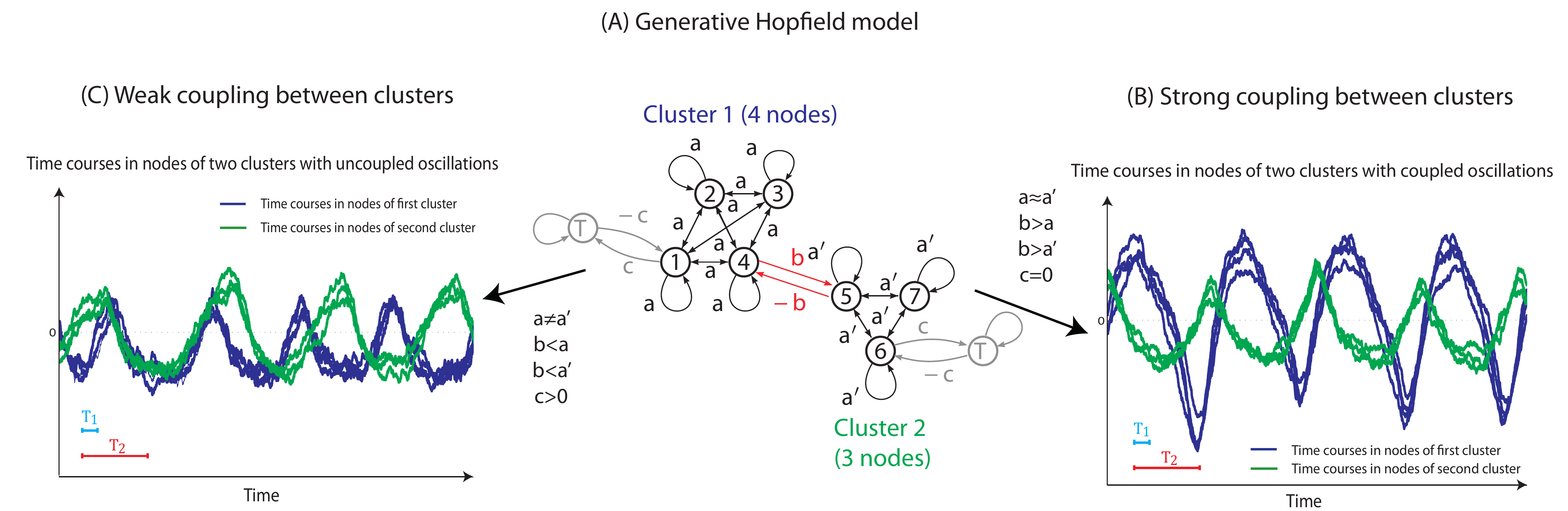}
      \caption{(A) The generative model contains two clusters and the values of the directed connectivity between the nodes is indicated on the corresponding arrows. These values correspond to the matrix C in (\ref{Hopfield}). (B) Set of parameters leading to coupled oscillations and (C) decoupled oscillations.}
      \label{model}
   \end{figure*}
\noindent where $x_i$ denotes the level of activity of the variable $i$, $G_{ij}$ is the strength of the connection from $x_j$ to $x_i$, and $sat$ is a sigmo\"idal saturation function. Fig. \ref{model} shows how we generate oscillations in two clusters. In the first case (Fig. \ref{model}B) the oscillations are coupled, leading to dephased oscillations of the same frequency whereas in the second case the clusters are decoupled (Fig. \ref{model}C), leading to oscillations of different frequencies in the two clusters. We finally generate three different datasets for each configuration by sampling these time series at different frequencies. The first dataset is produced by using the original time series (no sampling), the second and third datasets are obtained by sampling the time series with a period of T$1$ and T$2$ to generate synthetic data with higher frequency content.

\begin{figure}[thpb]
      \centering     
      \includegraphics[width=0.5\textwidth]{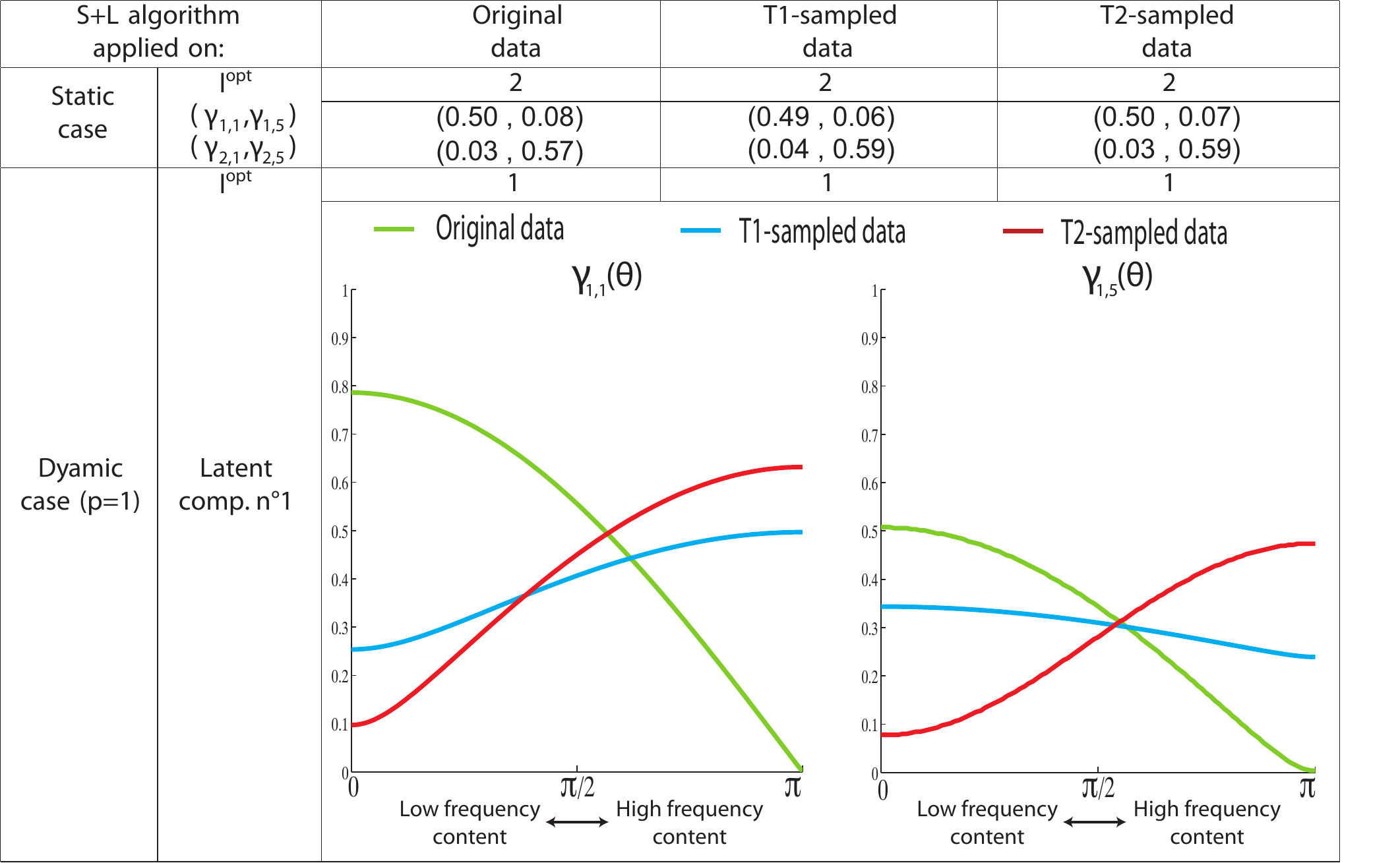}
      \caption{Latent components identified in the static and dynamic cases from the three datasets generated from coupled oscillators (Fig. \ref{model} B).}
      \label{same_f}
   \end{figure}

Fig. \ref{same_f} shows the static and dynamic latent components as defined in section \ref{lat_comp} that are identified in the optimal models from the synchronous oscillations datasets of Fig. \ref{model}B. Since the results are very similar within the nodes of each cluster and for clarity purposes we plot only the value of the latent components in node $1$ ($\gamma_{i,1}$) and in node $5$ ($\gamma_{i,5}$).
   
In the static case two latent components corresponding to the two clusters of the generative model are identified. There is no significant difference for the three input datasets suggesting that the frequency content of original data is not encoded in the static latent components. In the dynamic case ($p=1$), the latent components are encoded in $\gamma_{i,j}(\theta)$ and a single latent component is identified in the optimal model. Interestingly, the T2-sampled synthetic data (high-frequency time series) leads to a latent component showing a strong high-frequency content whereas the original dataset leads to a latent component with dominant low-frequency content.

      \begin{figure}[thpb]
      \centering
     
      \includegraphics[width=0.5\textwidth]{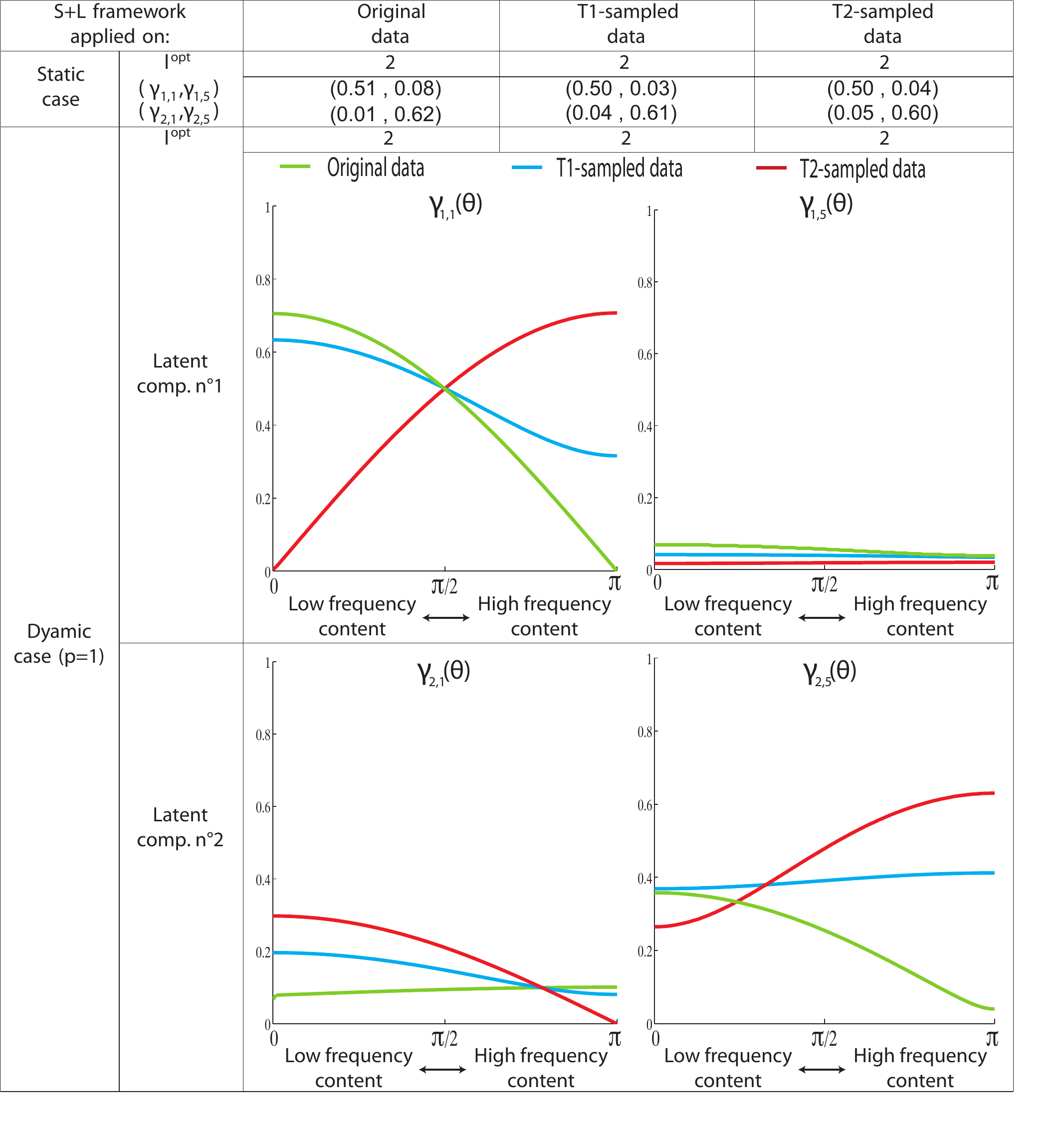}
      \caption{Latent components identified in the static and dynamic cases from the three datasets generated from decoupled oscillators (Fig. \ref{model}C).}
      \label{diff_f}
   \end{figure}
      
Having the same approach on asynchronous oscillations, we get the results shown in Fig. \ref{diff_f}. In the static case we still obtain two latent components recovering the two clusters with no distinction between the three starting datasets. In the dynamic case $(p=1)$, however, the optimal model now has two latent components, each one capturing the oscillations in one of the two clusters. As in the previous case the frequency content of the synthetic data is encoded in the frequency content of the latent component. Identifying two latent components probably comes from the fact that the frequency of oscillation in the two clusters are different, suggesting that the dynamic latent component identifies a \emph{spatio-temporal} subspace of variation common to different manifest variables.

\subsection{Application to neuroimaging data} 

\begin{figure}[thpb]
      \centering     
      \includegraphics[width=0.48\textwidth]{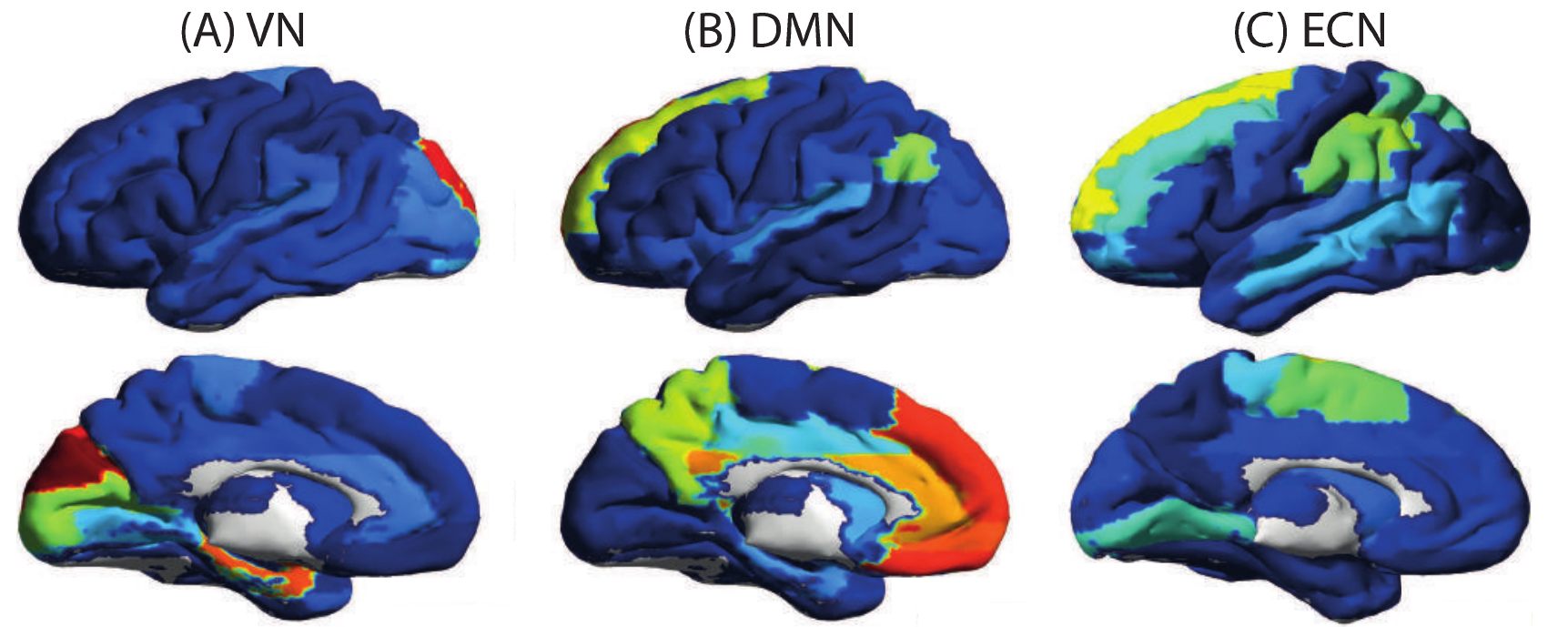}
      \caption{Three resting state neuronal networks are commonly recovered using component analysis: (A) the visual network, (B) the default mode network, and (C) the executive control network.}
      \label{3net}
   \end{figure}

We illustrate the application of our proposed algorithm on real neuroimaging data consisting of functional magnetic resonance imaging time series in 90 brain regions collected on 17 patients during rest \cite{vanh}. It should be noted that this problem dimension is not tractable with a standard optimization tool such as the CVX toolbox of \cite{grant}. The classical approach is to use component analysis to extract neuronal networks. During rest, three networks are robustly identified using component analysis: the visual network (VN), the default mode network (DMN) and the executive control network (ECN) that are represented in Fig \ref{3net}. 

\begin{figure}[thpb]
      \centering     
      \includegraphics[width=0.48\textwidth]{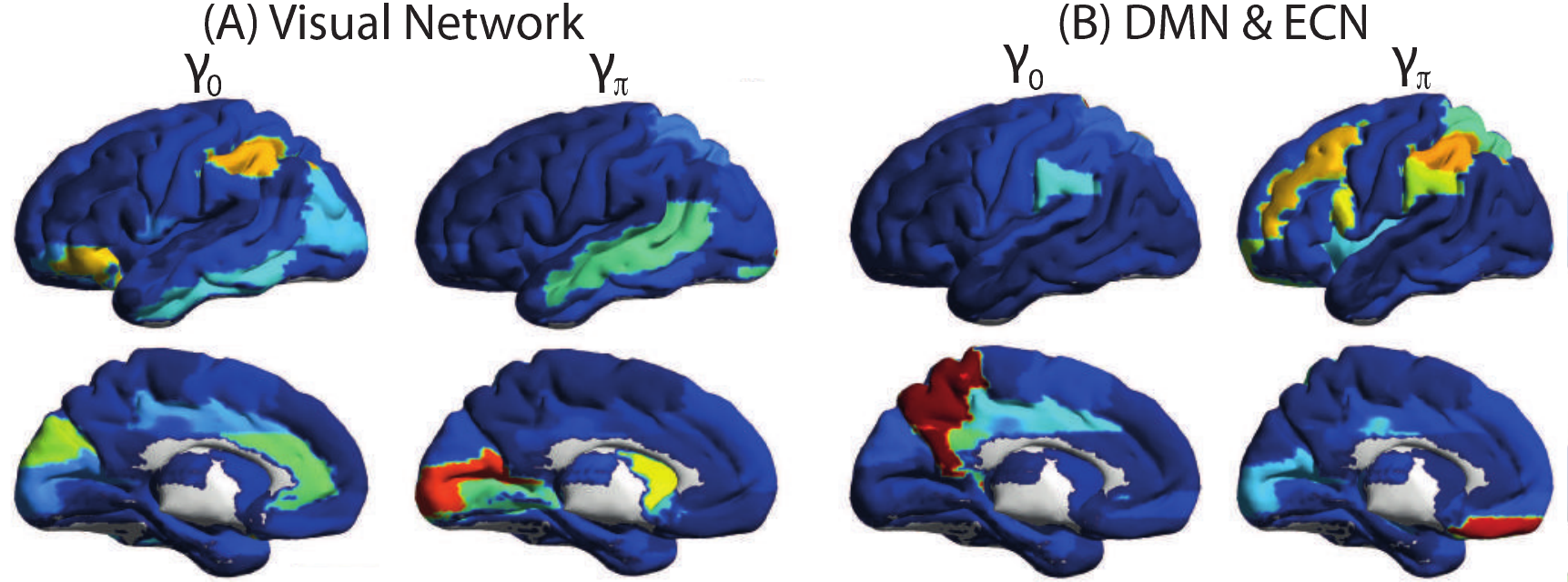}
      \caption{(A) The visual network is recovered in 14 out of 17 subjects. (B) The DMN and ECN are coupled into a unique latent component in 12 out of 17 subjects.}
      \label{dyn_net}
   \end{figure}

By fitting a first order AR model ($p=1$) to this dataset we identify latent components corresponding to these networks. In Fig. \ref{dyn_net} we plot only these components for $\theta=0$, a low-frequency contribution called $\gamma_0=\gamma(\theta=0)$ from the definition of Section \ref{lat_comp}, and $\theta=\pi$, a high-frequency contribution called $\gamma_{\pi}=\gamma(\theta=\pi)$. Indeed, for a first order model the latent components are characterized by these two extreme values $\gamma_0$ and $\gamma_{\pi}$.

In 14 subjects out of 17 we observe that VN is recovered in one latent component, with no significant differences between $\gamma_0$ and $\gamma_{\pi}$. On the other hand, in 12 subjects the DMN and ECN networks are gathered in a unique latent variable, the DMN corresponding to $\gamma_0$ and ECN to $\gamma_{\pi}$. This interestingly echoes some recent results suggesting that the DMN and the ECN are both consciousness related processes and are anti-correlated, oscillating at similar frequencies \cite{vanh}.

\section{Conclusion} 
The contribution of our work is twofold. First, we reformulated the sparse plus low-rank autoregressive identification problem into the ADMM framework in order to scale it to larger datasets encountered in neuroimaging applications. Second, we presented a deeper exploration of the type of information encoded in the latent components identified in the low-rank part of the decomposition. Fig. \ref{same_f} \& \ref{diff_f} suggest that this information is richer in dynamic models than in static models in the sense that dynamic latent components recover \emph{spatio-temporal} properties of the original time series such as common spectral content. Applied to a neuroimaging dataset, this interpretation led to a novel characterization of the dynamical interplay between two neuronal networks mediating consciousness, echoing recent experimental results.

As a future research direction we intend to explore whether additional information can be recovered in higher order dynamical latent components ($p>1$). It would also be interesting to study the mathematical link between the latent components of the proposed framework and the widely used static principal components that also capture a `common subspace' of variation of many variables from the covariance matrix instead of the precision matrix. Finally, in addition to using separability into sparse and low-rank constraints, it could be beneficial to exploit these structures in the algorithm updates in order to scale it to even larger problems.

\end{document}